\documentclass[letterpaper, 10 pt, conference]{ieeeconf}

\IEEEoverridecommandlockouts

\usepackage{graphicx}
\usepackage{amsmath}
\usepackage{amssymb}

\usepackage{cleveref}
\usepackage{multirow}
\usepackage{pifont}
\usepackage{booktabs}
\usepackage{xcolor}
\makeatletter
\def\fnum@table{Table~\thetable}
\let\IEEE@origmakecaption\@makecaption
\long\def\@makecaption#1#2{
  \ifx\@captype\@IEEEtablestring
    \setbox\@tempboxa\hbox{\footnotesize #1.~~ }
    \parbox[t]{\hsize}{\footnotesize\noindent\unhbox\@tempboxa #2}
    \@IEEEtablecaptionsepspace
  \else
    \IEEE@origmakecaption{#1}{#2}
  \fi}
\makeatother
\definecolor{figblue}{RGB}{25, 110, 228}
\definecolor{figred}{RGB}{253, 0, 57}
\definecolor{figfaded}{RGB}{230, 110, 140}
\definecolor{figpurple}{RGB}{203, 48, 224}
\definecolor{figgreen}{RGB}{60, 160, 90}
\usepackage{makecell}
\usepackage{tikz}
\usepackage{pgfplots}
\pgfplotsset{compat=1.18}
\usepackage{siunitx}
\usepgfplotslibrary{groupplots}
\pgfplotsset{compat=1.18}

\title{\LARGE \bf
Comfortably Stuck: Evaluation Pitfalls in Bench2Drive's Closed-Loop Protocol
}

\author{Emad Aghajanzadeh$^{1,2,*}$, Thomas Schwarzfischer$^{1,2,*,\dagger}$, Moritz Thoma$^{1,2}$, Nael Fasfous$^{1,\dagger}$,\\
Manoj Rohit Vemparala$^{1}$, Pierpaolo Mori$^{1}$, Daniel M\"uller-Gritschneder$^{3}$, and Ulf Schlichtmann$^{2}$
\thanks{This work has been submitted to the IEEE for possible publication. Copyright may be transferred without notice, after which this version may no longer be accessible.}
\thanks{$^{*}$Equal contribution. $^{\dagger}$Affiliated with BMW Group at the time the work was done.}
\thanks{$^{1}$BMW Group, Munich, Germany; $^{2}$Technical University of Munich (TUM), Munich, Germany; $^{3}$Vienna University of Technology (TU Wien), Vienna, Austria. Corresponding author: {\tt\small emad.aghajanzadeh@bmw.de}}
}

\begin{document}

\maketitle
\thispagestyle{empty}
\pagestyle{empty}

\begin{abstract}
Closed-loop evaluation is widely regarded as a more faithful way to assess end-to-end (E2E) driving models than open-loop evaluation, and Bench2Drive, built on the CARLA simulator, has become the dominant benchmark for it. We show that its evaluation protocol itself contains pitfalls that can misrepresent the models scored on it. Bench2Drive's comfort metric, adapted from nuPlan, has several implementation errors and a segmentation scheme that rewards standing still, so it mostly measures how long a vehicle stays stationary, correlating with stationary time at $r=0.97$ across eight models. We propose a corrected, distance-based metric that cuts that correlation to $r=0.06$. Under it, the strong driving-score/comfort tradeoff the original metric reports is no longer detectable once models are compared at matched driving speed, which suggests that comfort trades against how fast a model drives rather than against how well it drives. We also uncover two pitfalls in how the harness supplies navigation signals: the default controller can replace a model's own steering with an angle toward the route target point, and the target point supplied at evaluation time can differ substantially from what a model saw in training. We demonstrate both on UniAD and VAD, where their effects differ sharply, and argue that future closed-loop evaluations should check for them. We will release all fixes as standalone code, so corrected comfort scores can be obtained from logged trajectories without re-running closed-loop simulation.
\end{abstract}

\section{INTRODUCTION}

End-to-end (E2E) autonomous driving models have recently attracted substantial attention, and as the space grows, reliable evaluation becomes essential for directing research and comparing models fairly~\cite{dauner2024navsim, jia2024bench2drive}. Closed-loop evaluation, in which a model reacts to the consequences of its own past decisions rather than replaying a fixed expert trajectory, is a more faithful proxy of real-world driving quality than open-loop evaluation~\cite{renz2025simlingo,jia2024bench2drive}. CARLA~\cite{dosovitskiy2017carla} is the most widely used platform for closed-loop testing, and Bench2Drive~\cite{jia2024bench2drive} builds on it with a curated set of edge-case routes to standardize closed-loop evaluation across the field. We show that the Bench2Drive protocol itself contains issues that can lead to large misinterpretations of E2E models along two axes: driving comfort and driving behavior.

\textbf{Comfort:} Bench2Drive adapts nuPlan's comfort check~\cite{caesar2021nuplan}, which tests six kinematic quantities, among them lateral and longitudinal acceleration, jerk and yaw rate, against thresholds calibrated on human driving logged at $10\,\mathrm{Hz}$ with angles in radians. Ported to CARLA's $20\,\mathrm{Hz}$ logs, the implementation gets the sampling rate, the smoothing window, the angular units and the yaw derivative wrong, and together these errors inflate the measured jerk and yaw terms enough that segments in which the vehicle is genuinely driving fail thresholds a human driver would pass. Worse, comfort is scored over fixed one-second chunks rather than distance travelled, so a vehicle that simply stops satisfies every threshold for free and keeps banking comfortable segments for as long as it stands there. The resulting score measures how long a model stands still rather than how smoothly it drives, correlating with stationary time at $r=0.97$ across eight models and ranking the models that most often get stuck as the most comfortable. Fixing the implementation alone does not help, because time-based segmentation rewards standing still by construction. We therefore propose a distance-based metric that scores comfort per meter driven, which cuts that correlation to $r=0.06$ and moves VAD and SimLingo sharply down the comfort ranking. Measured this way, the strong tradeoff between driving score and comfort that prior evaluations report is no longer detectable once models are compared at matched driving speed, so comfort trades against how fast a model drives rather than against how well it drives.

\textbf{Navigation signals:} E2E models rely on a high-level navigation signal, a discrete command or a continuous target point, to know where the route leads. Prior work studies how best to integrate that signal into the architecture~\cite{nguyen2026lead, wu2025navihydra}; we instead study how the harness supplies it, and find two issues. First, Bench2Drive's default controller replaces the model's own predicted steering with an angle toward the route target point, a behavior we call steering overruling (SO), in more than half of all frames in some scenarios. This masks whether a model can correct its own drift or react to a sudden obstacle, and it affects models so unevenly that scores obtained under the default controller are not comparable across models. Second, the target point supplied at evaluation time can differ substantially in density and distance from what a model saw during training, pushing it toward out-of-distribution inputs. We propose fixes for both: a controller that never overwrites the model's steering, and a route planner whose target-point sampling matches the training distribution.

Our contributions are as follows:
\begin{enumerate}
    \item We show that Bench2Drive's comfort metric is dominated by a bias toward standing still, and propose a corrected, distance-based metric that removes it while staying faithful to nuPlan's thresholds.
    \item We uncover two protocol-level pitfalls in how Bench2Drive supplies navigation signals, a controller that overwrites the model's steering and a train-deployment mismatch in the target point, and propose fixes for both.
    \item We re-benchmark comfort across eight recent E2E models under the corrected metric and demonstrate the navigation fixes on UniAD and VAD. We will release all fixes as standalone code, so corrected comfort scores can be obtained from existing logs without re-running closed-loop simulation.
\end{enumerate}

\section{RELATED WORK}

\subsection{End-to-end driving architectures}
End-to-end (E2E) driving replaces the hand-designed interfaces of the modular stack with a single network trained toward the final planning task~\cite{zhao2025survey, hu2023planning}. In E2E design, modules exchange information with dense features which preserves scene information and limits error accumulation~\cite{yang2025survey}. UniAD~\cite{hu2023planning} and VAD~\cite{jiang2023vad} pioneered this direction. Both chain perception, prediction, and planning through task queries over a shared bird's eye view (BEV) representation, and VAD swaps the costly rasterized scene for a vectorized one. Transfuser~\cite{chitta2022transfuser} fuses camera and LiDAR modalities into a dense BEV with transformer attention, and TFv6~\cite{nguyen2026lead} is the latest variant of it that further enhances its planning head design. Models also differ in how the plan itself is produced. Hydra-MDP~\cite{li2024hydra} and SparseDriveV2~\cite{sun2026sparsedrivev2} score a fixed trajectory vocabulary instead of regressing a single trajectory, the latter from a sparse scene representation. SimLingo~\cite{renz2025simlingo} and Orion~\cite{fu2025orion} put a language model in the loop: SimLingo outputs reasoning and action directly from a Qwen model~\cite{yang2025qwen3}, while Orion decodes language reasoning features into actions with a generative planner.

\subsection{Evaluation of End-to-end models}
\paragraph{Closed-loop benchmarks}

E2E models are evaluated either open-loop or closed-loop~\cite{dauner2024navsim, cao2025pseudo, jia2024bench2drive}. Open-loop evaluation replays pre-recorded expert data and measures how closely the model imitates it. It is cheap and needs no simulator, but it never confronts the model with the consequences of its own actions. Real driving is closed-loop: every scene a driver meets follows from what it did before, including its errors. Closed-loop evaluation is therefore the more faithful measure of driving quality, and safety confines it to simulation. Real data supports it only partially. NAVSIM~\cite{dauner2024navsim} rolls a predicted trajectory out for four seconds against replayed actors, and NAVSIMv2~\cite{cao2025pseudo} adds perturbed restarts to probe recovery, but both query the policy once and score a plan it never revises. HUGSIM~\cite{zhou2025hugsim} goes further, reconstructing recorded scenes with Gaussian splatting so the ego can leave the logged path and still be rendered from novel views, yet its HD-Score is inherited from NAVSIM and still scores the predicted trajectory rather than the path actually driven~\cite{xiong2026terratransfer}.
CARLA~\cite{dosovitskiy2017carla} closes the loop fully. The simulator reacts to the policy at every step, and renders the future scene sensors necessary for E2E models, different than planning-only benchmarks like NuPlan~\cite{caesar2021nuplan}. Bench2Drive~\cite{jia2024bench2drive} design 220 driving routes, each built around one edge-case scenario and one driving ability, scored by driving score, success rate, efficiency, and comfort. It is now the default closed-loop benchmark for E2E driving~\cite{jia2025drivetransformer, sun2026sparsedrivev2, fu2025orion, renz2025simlingo, li2025hydra}.

\paragraph{Navigation signals in closed-loop evaluation}
E2E policies need a high-level navigation signal to know where the route leads. It arrives either as a discrete navigation command~\cite{hu2023planning, jiang2023vad, fu2025orion} or as a continuous navigation target point from a route planner~\cite{renz2025simlingo, nguyen2026lead}. Jaeger et al.~\cite{jaeger2023hidden} show that target-point-conditioned models learn to follow the point as a shortcut for lateral control, so gains credited to architecture partly reflect that bias~\cite{nguyen2026lead}. That bias is acquired from training data. We study the evaluation side instead. First, the harness supplies its own target point at test time, and its distribution need not match the training one. Fixing the distance to a constant does not resolve this: the point lies ahead on the route, so its distance shrinks as the ego approaches and jumps when the next point activates. What the harness controls is route sampling density, and every spacing induces a different distance distribution. Second, the harness does more than pass the signal on. Bench2Drive's controller steers toward the target point itself, overruling the model (\cref{sec:steering_issue}).

\paragraph{Comfort metrics}

nuPlan~\cite{caesar2021nuplan} defines an ego comfort check over six kinematic quantities: longitudinal and lateral acceleration, yaw rate, yaw acceleration, longitudinal jerk, and jerk magnitude. It is all-or-nothing. Every quantity must stay within its threshold on every frame, so one violation fails the whole trajectory. The thresholds~\cite{nuplan_devkit} are fit empirically on expert driving, given in SI units with angles in radians, and calibrated for a $10\,\mathrm{Hz}$ sampling rate together with the smoothing windows applied at that rate.
The formulation spread beyond nuPlan. NAVSIM~\cite{dauner2024navsim} scores comfort against the same style of thresholds, following PDM-Closed's weighting scheme~\cite{dauner2023parting}, and Bench2Drive~\cite{jia2024bench2drive} adopts them for CARLA. Comfort is thus shared across open-loop, pseudo-closed-loop, and closed-loop protocols rather than specific to any one of them.

Human-perception studies justify the ingredients. de Winkel et al.~\cite{de2023standards} tie discomfort to acceleration and jerk, and show it depends on the direction of motion as well as its magnitude. Hu et al.~\cite{hu2025autonomous} score comfort from jerk on Bézier-interpolated trajectories and add energy consumption. None of this work asks whether such a check still measures what it was calibrated for once it is ported into a closed-loop harness. We show that it does not: implementation details and the choice of segmentation dominate the score.

\section{EVALUATION PITFALLS AND CORRECTIONS}

\begin{figure}[!htpb]
    \centering
    \includegraphics[width=0.9\columnwidth]{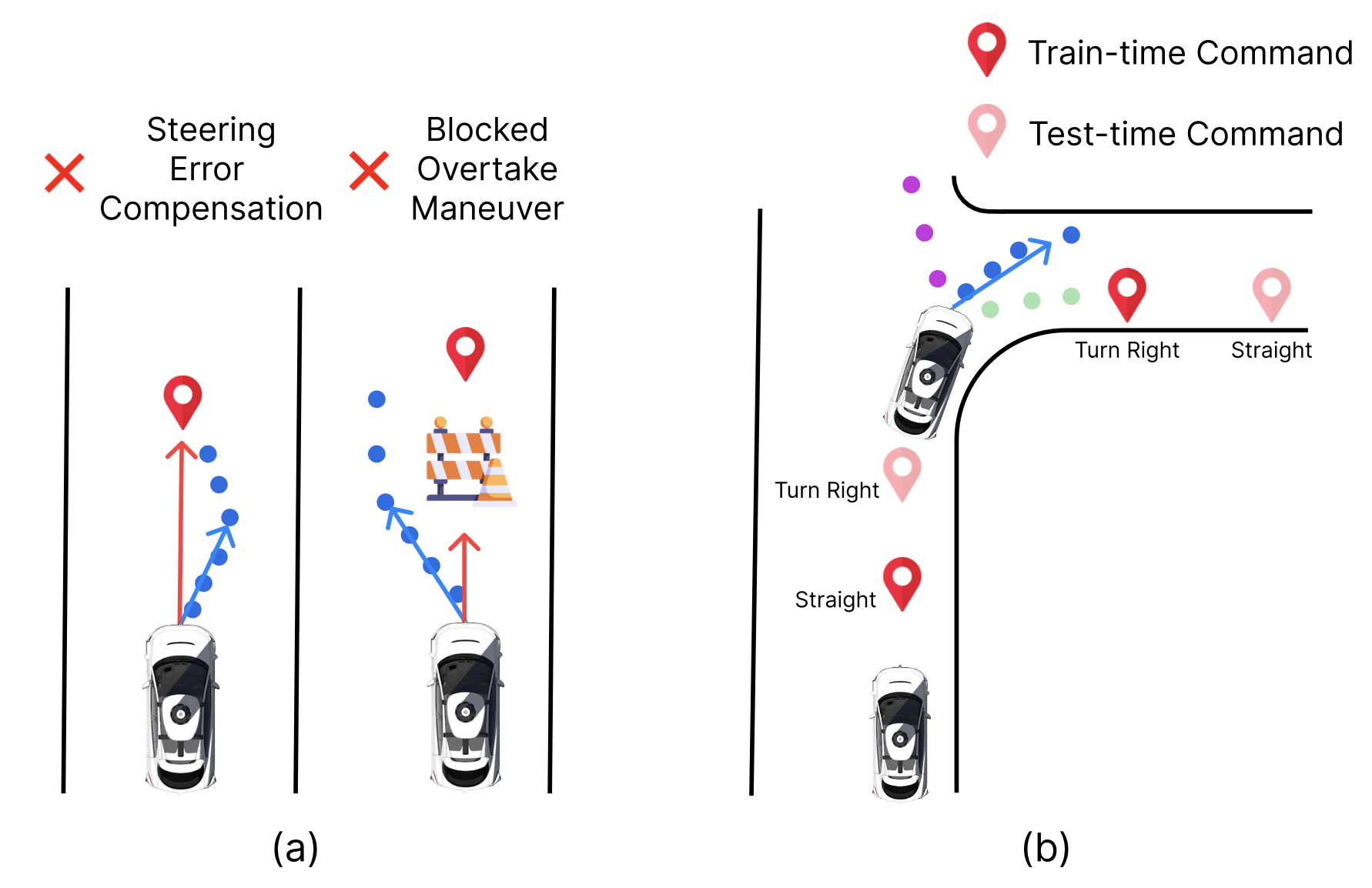}
    \caption{Illustration of two closed-loop evaluation pitfalls. In both panels \textcolor{figblue}{blue} marks the model's own output: its predicted waypoints and the steering vector they imply. (a) Steering overruling: the \textcolor{figred}{red} vector is the steering the controller substitutes, aimed at the navigation target point (\textcolor{figred}{red} pin). This can mask a model's failure to compensate its own steering error (left) and can override a maneuver around an obstacle (right). (b) Target-point mismatch: because the evaluation route planner samples the route differently than the planner that collected the training data, the navigation command switches at a different place along the route at test time (\textcolor{figfaded}{faded red} pins) than during training (\textcolor{figred}{red} pins). Here the ego receives \emph{turn right} early and begins to rotate, then receives \emph{straight} at the junction: the command calls for the straight continuation (\textcolor{figpurple}{purple}) while its heading and the road geometry support completing the right turn (\textcolor{figgreen}{green}), a command-context combination absent from training. Interpolating between the two can produce dangerous behavior. SO: steering overruling; TPM: target point mismatch.}
    \label{fig:cl_issues_image}
\end{figure}

We present these pitfalls in the order they occur within a closed-loop episode: navigation-related issues, which arise during decision-making and route-following, are discussed first (\cref{sec:steering_issue,sec:tp_mismatch_issue}) and illustrated in \cref{fig:cl_issues_image}, followed by the comfort metric, which is computed post-hoc from the logged trajectory (\cref{sec:comfort_issues}).

\subsection{Steering Overruling in B2D's Controller (SO)}
\label{sec:steering_issue}
During training, E2E models are typically supervised with expert trajectories rather than low-level control actions, since trajectories more consistently reflect the planned action and are less noisy. Most E2E models therefore predict a trajectory and need a controller to turn it into control actions. By default, B2D uses a PID controller that converts the predicted waypoints into steering for lateral control and throttle or brake for longitudinal control.

We find a subtle yet detrimental issue in the steering behavior of this default controller, which we call steering overruling (SO). B2D computes three candidate steering angles; the two that matter here are angle, toward an aim point taken from the model's own waypoints at about \SI{4}{\meter}, and angle\_target, toward the route target point. B2D selects angle\_target whenever it is smaller than angle or the tracking error is large, which is intended to reduce noise on straight roads and help with sudden turn commands. The condition triggers far more often than a ``subtle stabilization'' mechanism would warrant, however, overwriting the model's own steering decision in more than half of all frames in some scenarios. Models are then never exposed to their own compounding prediction error, one of the central properties closed-loop evaluation exists to capture, and the substitution can suppress evasive maneuvers, since the sharp angle needed to swerve around an obstacle is replaced by the shallower angle toward the target point.

We evaluate the impact of SO by disabling it and rerunning UniAD and VAD on Dev10, Bench2Drive's 10-route development subset with one route per scenario category, curated by DriveTransformer~\cite{jia2025drivetransformer} and now common practice. We also report a constant-velocity baseline: it ignores the sensor input entirely, drives at a constant \SI{5}{\meter\per\second}, and steers only toward the route target point, so what it scores reflects route-following alone. These results are reported together with our target-point fix from \cref{sec:tp_mismatch_issue} in \cref{tab:nav_issues}.

The effect of removing SO is highly asymmetric. It collapses UniAD's driving score from 53.42 to 19.76 and its success rate to zero, while VAD is essentially unchanged (52.37 to 52.49). A large share of UniAD's reported performance therefore came from the controller steering on its behalf rather than from the model itself. Our constant-velocity baseline puts this in context: it scores 48.57, within five points of both models under the default protocol. Dev10's driving score therefore barely separates two of the field's most-cited baselines from route-following alone.

Because the same default controller masks different models by very different amounts, Bench2Drive scores obtained under it are not comparable across models. We recommend that the controller not be permitted to overwrite the model's steering at all. For results already reported under the default controller, the SO-removed score should be reported next to them, since the default number on its own does not say how much of the driving the model did.

\subsection{Target Point Mismatch}
\label{sec:tp_mismatch_issue}
Another often overlooked aspect of E2E models' evaluation is how the navigation signal is supplied. These models require such a signal to know where the designed route leads, and in recent models it takes one of two forms: a navigation target point, given as continuous $(x,y)$ coordinates relative to the ego, or a navigation command, a discrete instruction such as turn left. While prior work studies how best to integrate the navigation signal into the model architecture~\cite{wu2025navihydra, nguyen2026lead}, we instead identify a train-deployment discrepancy in how that signal is produced by the harness itself.

\paragraph{How the navigation target is produced}
A CARLA route is specified as sparse keypoints, which the leaderboard interpolates into a dense polyline at roughly \SI{1}{\meter} spacing, annotating each point with a discrete road option. At every tick the harness emits one navigation target drawn from this polyline: an $(x,y)$ point in the ego frame together with the road option attached to that same point. The two forms of navigation signal are therefore not independent. Both are read from one polyline index, so whatever governs which index is emitted governs both the target point a model receives and the command it receives. Their value is set by two choices in the harness: which representation of the route is traversed, and when the index into it advances.

\paragraph{Where collection and evaluation diverge}
Both choices differ between data collection and evaluation. In the training data, the target is held at a near-constant short distance ahead of the ego, with a median of \SI{4.94}{\meter} and \SI{90}{\percent} of frames falling between \SI{4.06}{\meter} and \SI{6.60}{\meter}, and it moves smoothly: fewer than \SI{4}{\percent} of frames show a change above \SI{1.5}{\meter}. The original evaluation planner instead traverses the route after downsampling it, selecting the next waypoint outside a \SI{4}{\meter} radius. Because that downsampled plan retains only a small fraction of the interpolated points, the emitted target sits roughly an order of magnitude farther ahead and advances in discontinuous jumps of tens of meters. We quantify this gap for UniAD and VAD in \cref{fig:tp_dist}: a model trained under the former regime and evaluated under the latter is pushed toward inputs it never saw, in both the point it is given and the command that travels with it.

\paragraph{An aligned route planner}
Our planner addresses both choices. We traverse the dense polyline rather than a downsampled copy, and track it with a cursor that only ever moves forward along the route, so the target progresses continuously instead of jumping. This alone recovers the temporal behavior but not the geometry: advancing as soon as the ego passes the current target imposes no look-ahead, and the target collapses onto the ego. We therefore advance the cursor while the candidate point lies behind the ego or nearer than a standoff $R$, stopping at the first point beyond $R$. Advancing in a loop rather than by one index per tick makes the standoff independent of the control frequency, which itself differs between collection and evaluation. We set $R = \SI{4.06}{\meter}$, the 5th percentile of the training-time target distance. This value comes from training statistics alone; it was fixed before evaluation and never tuned against driving score.

\paragraph{Where the alignment remains incomplete}
The aligned planner corrects the median for both models but not the whole distribution. For UniAD it also fixes the tail, with p95 falling from \SI{130.2}{\meter} to \SI{20.0}{\meter} and frames whose target lies behind the ego dropping from \SI{13.7}{\percent} to \SI{0.9}{\percent}. For VAD neither improves: p95 moves only from \SI{48.8}{\meter} to \SI{45.6}{\meter}, and behind-ego frames rise from \SI{0.6}{\percent} to \SI{12.7}{\percent}. These are the same frames, all carrying a target \SIrange{44}{53}{\meter} away, so they are not the cursor collapsing onto the ego but frames in which the ego has left the intended path and the route runs back behind it. The mismatch is therefore only partly resolved for VAD.

Both models nonetheless improve on top of the SO fix, UniAD from 19.76 to 26.06 and VAD from 52.49 to 59.32 (\cref{tab:nav_issues}), so target-point sampling in closed-loop should match the sampling used when the training data was collected. The two end in very different places. VAD rises above its default score, so its reported standing survives the corrections, while UniAD reaches roughly half of what the default protocol credits it with, suggesting its Bench2Drive standing is largely an artifact of the harness. While probing UniAD we noticed that it solves perception and prediction well while its planning head makes little use of those features, which may explain the gap.

\begin{table}[t]
  \centering
  \small
  \setlength{\tabcolsep}{4pt}
  \caption{Cumulative impact of removing steering overruling (SO) and aligning target-point sampling (TPM) on driving score (DS) and success rate (SR), on Dev10, against a constant-velocity baseline. $\checkmark$ marks a fix applied. TPM is applied on top of SO, not independently.}
  \label{tab:nav_issues}
  \begin{tabular}{lccrr}
    \toprule
    \textbf{Model} & \textbf{SO} & \textbf{TPM} & \textbf{DS} & \textbf{SR (\%)} \\
    \midrule
    Const.\ velocity & $\times$ & -- & 48.57 & 20.0 \\
    \midrule
    \multirow{3}{*}{UniAD} & $\times$     & $\times$     & 53.42 & 30.0 \\
                           & $\checkmark$ & $\times$     & 19.76 & 0.0 \\
                           & $\checkmark$ & $\checkmark$ & 26.06 & 10.0 \\
    \midrule
    \multirow{3}{*}{VAD}   & $\times$     & $\times$     & 52.37 & 30.0 \\
                           & $\checkmark$ & $\times$     & 52.49 & 30.0 \\
                           & $\checkmark$ & $\checkmark$ & 59.32 & 30.0 \\
    \bottomrule
  \end{tabular}
\end{table}

\begin{figure}[t]
    \centering
    \definecolor{tpTrain}{HTML}{0072B2}
\definecolor{tpOrig}{HTML}{D55E00}
\definecolor{tpOurs}{HTML}{009E73}

\resizebox{\linewidth}{!}{

\begin{tikzpicture}[
    /pgfplots/every axis/.append style={
      scale only axis,
      width=0.84\columnwidth,
      grid=both,
      grid style={gray!18, line width=0.2pt},
      major grid style={gray!30, line width=0.25pt},
      tick label style={font=\scriptsize},
      label style={font=\scriptsize},
      title style={font=\scriptsize},
      tick align=outside,
      legend cell align={left},
      legend style={font=\scriptsize, draw=gray!60, fill=white,
                    fill opacity=0.92, text opacity=1,
                    inner sep=2pt, row sep=-1.5pt},
    },
  ]

  \begin{axis}[
    scale only axis, width=0.84\columnwidth,
    height=0.32\columnwidth,
    xmode=log,
    xlabel={target-point distance [m]},
    ylabel={density},
    title={Marginal distribution of target-point distance (UniAD)},
    xmin=1.5, xmax=160,
    ymin=0, ymax=0.55,
    xtick={2,5,10,20,50,100},
    xticklabels={2,5,10,20,50,100},
    ytick={0,0.1,0.2,0.3,0.4,0.5},
    legend style={at={(0.98,0.96)}, anchor=north east, legend columns=1},
  ]

    \addplot[draw=none, fill=tpTrain, fill opacity=0.18, forget plot]
      coordinates {(4.06,0) (4.06,0.55) (6.60,0.55) (6.60,0)} \closedcycle;
    \addlegendimage{area legend, draw=none, fill=tpTrain, fill opacity=0.18}
    \addlegendentry{Training p5-p95}

    \addplot[const plot, no marks, thick, color=tpTrain]
      table[x=bin_left, y=density, col sep=comma] {figures/tp_dist/uniad_train_hist.csv};
    \addlegendentry{Training}

    \addplot[const plot, no marks, thick, color=tpOrig]
      table[x=bin_left, y=density, col sep=comma] {figures/tp_dist/uniad_buggy_hist.csv};
    \addlegendentry{Original planner}

    \addplot[const plot, no marks, thick, color=tpOurs]
      table[x=bin_left, y=density, col sep=comma] {figures/tp_dist/uniad_newchange_hist.csv};
    \addlegendentry{Our planner}

  \end{axis}

\end{tikzpicture}
}

    \vspace{3pt}
    \scriptsize
    \begin{tabular}{lcccc}
        \toprule
        \multirow{2}{*}{Model} & \multicolumn{2}{c}{Original planner} & \multicolumn{2}{c}{Our planner} \\
        & Median & p95 & Median & p95 \\
        \midrule
        UniAD & 64.60 & 130.16 & 6.28 & 20.03 \\
        VAD   & 16.64 & 48.82  & 6.09 & 45.64 \\
        \bottomrule
    \end{tabular}

    \caption{Target-point distance supplied to UniAD in training and at evaluation time, on a logarithmic axis, with statistics (\si{\meter}) for both models below. The shaded band is the training p5 to p95 range, \SIrange{4.06}{6.60}{\meter}, shared by both models. The original planner places UniAD's target at a median of \SI{64.6}{\meter}, over an order of magnitude beyond the \SI{4.94}{\meter} training median, and behind the ego in \SI{13.7}{\percent} of frames. Ours restores the median for both models, to \SI{6.28}{\meter} and \SI{6.09}{\meter}. For UniAD it also removes the tail (p95 \SI{130.2}{\meter} to \SI{20.0}{\meter}) and the behind-ego frames (\SI{13.7}{\percent} to \SI{0.9}{\percent}); for VAD it does not, and we report that residual in \cref{sec:tp_mismatch_issue}.}
    \label{fig:tp_dist}
\end{figure}

\subsection{Comfort}
\label{sec:comfort_issues}
While most papers optimize primarily for driving score, comfort is still widely reported and should be an important tool for characterizing the tradeoff between driving-score improvements and ride comfort. Unfortunately, we identify several issues with how this metric is calculated in Bench2Drive. We argue that the current implementation does not faithfully reflect driving comfort, and instead propose a revised metric that addresses these shortcomings.

\subsubsection{Bench2Drive's comfort metric}
Bench2Drive~\cite{jia2024bench2drive} follows nuPlan~\cite{caesar2021nuplan} in computing this metric by comparing six of the ego vehicle's kinematic signals (longitudinal and lateral acceleration, jerk magnitude, longitudinal jerk, yaw acceleration, and yaw rate) against thresholds calibrated on human driving data. Routes are split into 1-second segments, motion statistics are computed for each segment, and a segment is considered comfortable only if every frame in it stays within the defined thresholds. Route comfort is then the ratio of comfortable segments, and a model's final comfort score is the average of its route comfort scores. However, this official implementation has a few major issues.

\paragraph{Implementation errors}
First and most importantly, there are some implementation errors which invalidate the metric:
\begin{itemize}
    \item \textbf{Sampling-rate mismatch.} Some of these signals require a derivative, computed at an assumed frequency of $10\,\mathrm{Hz}$, whereas the CARLA simulator runs and logs at $20\,\mathrm{Hz}$. All jerks and the yaw acceleration are therefore scaled by a factor of two, inflating violations of the jerk bounds.

    \item \textbf{Under-sized smoothing window.} For smoothing and differentiating the raw CARLA signals, a single 7-frame window is used, far shorter than the effective support nuPlan uses for jerk. This shorter window amplifies high-frequency noise in the acceleration, producing larger jerk estimates than a human passenger would actually perceive. nuPlan's window size must be preserved if the resulting signals are to be thresholded against nuPlan's calibrated values.

    \item \textbf{Invalid yaw processing.} The ``yaw acceleration'' is obtained by smoothing the yaw rate \emph{without} taking a derivative, so it does not actually represent an acceleration, and comparing it against yaw-acceleration thresholds is unsound.

    \item \textbf{Mismatched yaw unit.} The thresholds for yaw acceleration and yaw rate in nuPlan are given in radians, while CARLA reports these quantities in degrees. Bench2Drive's implementation does not convert between the two, introducing an additional factor of $\sim57$ on these two signals.
\end{itemize}

\paragraph{Chunk-boundary artifacts}
To calculate the six comfort measures, Bench2Drive collects intermediate kinematic signals from CARLA: the acceleration, the forward and right unit vectors, the yaw rate, and the ego location. These signals are split into segments and smoothed and differentiated independently within each one. Because this happens in isolated 20-frame chunks, edge effects are introduced that distort the physical measurements and cause threshold violations even when the underlying motion is smooth.

\paragraph{Temporal segments reward standing still}
While nuPlan evaluates a trajectory over a short scenario, Bench2Drive's routes are considerably longer; nuPlan's all-or-nothing principle would make a comfort test over an entire route almost always fail~\cite{jia2024bench2drive}. Bench2Drive therefore splits each route into short, 20-frame segments and scores these individually. However, this introduces a critical bias: because the route is segmented into fixed 1-second intervals rather than by route progress, a stationary vehicle continues generating comfortable segments indefinitely, since no movement means every threshold is trivially satisfied. Meanwhile, due to the implementation errors above, segments where the vehicle is actually moving are almost always considered uncomfortable. As a result, the comfort metric mostly measures how much time a vehicle spends stationary rather than how comfortably it drives: models that perform worse on driving score and route completion almost always achieve a higher comfort score, since they tend to get stuck early and continue accumulating comfortable, stationary segments. This is compounded by the fact that a route where the model gets stuck early covers less distance than a completed one, yet contributes the same number of comfort segments, making the easy first few meters of a route disproportionately important to the final score.

\subsubsection{Proposed Metric}
\label{sec:proposed_comfort}

To address these issues, we propose a corrected formulation of the comfort metric that fixes all of the concerns above while staying as close as possible to nuPlan's and Bench2Drive's original thresholds and calculations. It rests on three changes: \emph{nuPlan-faithful} calculation, \emph{smooth-first} filtering, and \emph{distance-based} segmentation.

\paragraph{nuPlan-faithful calculation}
nuPlan operates at $10\,\mathrm{Hz}$, smoothing acceleration with an $8$-frame window and computing jerk over $15$ frames, i.e. time spans of $0.8\,\mathrm{s}$ and $1.5\,\mathrm{s}$ respectively. To preserve these physical time spans at CARLA's $20\,\mathrm{Hz}$, we double the window lengths to $w_a = 16$ and $w_j = 30$ frames. Together with the corrected $\Delta t$, this makes the smoothed accelerations and jerks directly comparable to the values against which nuPlan's thresholds were calibrated. The yaw rate is converted from CARLA's $\mathrm{deg/s}$ convention to $\mathrm{rad/s}$, and we apply an explicit derivative to extract the yaw acceleration from the yaw rate.

\paragraph{Smooth-first filtering}
Instead of smoothing each chunk in isolation, we apply the same Savitzky-Golay filter~\cite{savitzky1964smoothing} (polynomial order $2$, matching nuPlan's) \emph{once} over the entire episode and only afterwards slice the resulting signals into segments. This eliminates the chunk-boundary artifacts of the original metric, since no segmentation is visible to the filter. We also use the correct derivative time step, $\Delta t = 0.05\,\mathrm{s}$, matching CARLA's $20\,\mathrm{Hz}$ logging rate.

\paragraph{Distance-based segmentation}
Rather than temporal chunks, we partition each route into $N=100$ equidistant bins along the reference polyline, removing the bias introduced by segments where the vehicle is standing still. We project every ego location $\mathbf{p}_t$ onto the route to obtain its normalized progress and assign each frame to the corresponding bin. Each bin spans $1\%$ of the route length ($\sim\!1-2\,\mathrm{m}$) and is comfortable only if all of its frames are comfortable, similar to nuPlan's and Bench2Drive's binning. The choice of $N$ moves absolute scores considerably but not the ordering they induce: sweeping $N$ over the models for which we logged it, the ranking at $N=10$, $20$ and $50$ is identical to the one at $N=100$ (Spearman $\rho=1.00$), so comparisons between models do not depend on this choice.
As an additional measure, we aggregate at the segment level across all routes rather than averaging per-route scores. This gives each driven meter of road equal weight, so a route where the model gets stuck early no longer contributes a full route's worth of ``comfortable'' stationary bins, and a segment's importance is tied to distance traveled rather than time spent, including time spent standing still.

\section{RESULTS}
\label{sec:results}

\subsection{Evaluation Protocol}
\label{sec:eval_protocol}

We evaluate eight models on Bench2Drive's full 220-route benchmark: UniAD~\cite{hu2023planning}, VAD~\cite{jiang2023vad}, SparseDriveV2~\cite{sun2026sparsedrivev2}, TFv6~\cite{nguyen2026lead}, Orion~\cite{fu2025orion}, SimLingo~\cite{renz2025simlingo}, and two privileged models, CaRL~\cite{jaeger2025carl} and PDM-Lite~\cite{sima2024drivelm}. The set is deliberately broad, spanning language-based, generative and trajectory-scoring planners alongside reinforcement-learned and rule-based ones, so that our comfort analysis is not tied to a single family of designs. For each model we load the officially released checkpoint and re-run closed-loop evaluation ourselves, because all three comfort variants need the per-frame ego state (position, velocity, heading, and their derivatives) over the whole rollout, which no release retains or reports. That log is also the only thing our correction requires, since computing comfort from it is a pure recomputation (\cref{sec:comfort_issues}): we will release the evaluation harness together with the corrected comfort computation, so anyone who keeps the per-frame log can obtain corrected scores without re-running simulation. The two pitfalls have different scopes. Steering overruling can reach a model only if it hands its predicted waypoints to Bench2Drive's default controller, which is a property of the released code: two of our eight models do so, UniAD and VAD. For that pitfall we therefore correct every model in our set that it can affect. Target-point mismatch is not settled the same way, since whether a model is exposed depends on how its own training data was collected, which we did not audit for the remaining six. We demonstrate it on UniAD and VAD and treat it as a risk that other models may share rather than as a closed question.

\begin{table*}[t]
    \centering
    \caption{Closed-loop results on 220 Bench2Drive routes: the traditional comfort metric, a fixed implementation of it, and our distance-based metric. UniAD and VAD use the corrected protocol (SO removed, TPM fixed; \cref{sec:steering_issue,sec:tp_mismatch_issue}).}
    \label{tab:comfort_main}
    \begin{tabular}{c|ccc|c|ccc}
        \toprule
        Method &
        \makecell{Traditional\\Comfort} &
        \makecell{+ fixed\\implementation} &
        \makecell{+ distance-based\\segmentation (ours)} &
        \makecell{Time spent\\stationary (\%)} &
        Efficiency &
        \makecell{Success\\Rate (\%)} &
        \makecell{Driving\\Score} \\
        \midrule

        UniAD~\cite{hu2023planning} & 39.28 & 79.06 & 72.93 & 45.67 & 135.20 & 10.96 & 37.58 \\

        VAD~\cite{jiang2023vad} & 45.95 & 56.84 & 37.21 & 47.64 & 152.69 & 17.89 & 46.18 \\

        SparseDriveV2~\cite{sun2026sparsedrivev2} & 7.75 & 23.50 & 50.28 & 7.08 & 211.98 & 70.00 & 87.17 \\

        TFv6~\cite{nguyen2026lead} & 19.58 & 25.97 & 29.20 & 19.83 & 207.17 & 85.91 & 94.47 \\
        
        Orion~\cite{fu2025orion} & 14.76 & 70.24 & 75.98 & 23.07 & 151.69 & 51.14 & 75.33 \\
        
        SimLingo~\cite{renz2025simlingo} & 31.71 & 36.80 & 25.86 & 32.16 & 225.91 & 71.23 & 87.68 \\
        \midrule
        CaRL~\cite{jaeger2025carl} & 22.81 & 61.36 & 71.84 & 24.76 & 177.60 & 57.27 & 81.82 \\
        PDM-Lite~\cite{sima2024drivelm} & 22.77 & 32.57 & 35.82 & 20.43 & 250.80 & 91.19 & 96.59 \\
         \bottomrule
    \end{tabular}
\end{table*}

\subsection{Traditional Comfort Reveals a Measurement Bug}
\label{sec:comfort_bug}

We report efficiency, success rate, driving score, and the average time each model spends stationary alongside the three comfort variants. Stationary time is calculated using the same 1-second segments as the original Bench2Drive comfort metric, where a segment is considered stationary if the speed does not exceed \SI{0.1}{\meter\per\second}, the speed which Bench2Drive considers stuck. Comparing the \emph{Traditional Comfort} and \emph{Time spent stationary} columns of \cref{tab:comfort_main} shows a strong correlation between the traditional Bench2Drive comfort metric and the time a model spends stationary (Pearson correlation coefficient $r = 0.966
$).
This correlation stems from the fact that the existing metric is flawed, almost exclusively deeming stationary segments as comfortable. This behavior results in models with generally lower performance (e.g., UniAD, VAD) achieving higher ``comfort'' scores, since the metric rewards standing still and worse-performing models get stuck more often. Both are a case in point: they hold the two highest traditional comfort scores in \cref{tab:comfort_main} (39.28 and 45.95) while also spending the most time stationary (45.67\% and 47.64\%) and posting the two lowest driving scores.

\subsection{Corrected Comfort Metric}
\label{sec:corrected_comfort_results}

We next analyze comfort after addressing the implementation errors and chunk-boundary artifacts by smoothing before segmenting. We consider this a corrected implementation of the metric Bench2Drive describes. It can detect comfortable segments even while the vehicle is driving, which raises scores across the board, reported in the \emph{+ fixed implementation} column of \cref{tab:comfort_main}.

Fixing the implementation is not sufficient, because time-based segments keep accruing while the vehicle is not driving and a stationary vehicle satisfies every threshold trivially. Time-based segmentation therefore rewards standing still whether or not the implementation is correct. The table shows this directly: although the corrected implementation raises most models, the three that spend the most time stationary, UniAD, VAD and SimLingo, score \emph{lower} under our distance-based metric than under it, because part of what they had been credited with was time spent stopped. Distance-based segmentation removes the incentive rather than dampening it. Standing still advances no segment, so the score does not move at all while the vehicle is stopped, and comfort is measured over the distance actually driven.

\begin{figure}[t]
    \centering
    \resizebox{\linewidth}{!}{

\begin{tikzpicture}
\begin{axis}[
    width=\linewidth,
    height=0.58\linewidth,
    axis x line=bottom,
    axis y line=left,
    xlabel={Time [s]},
    ylabel={Speed [m/s]},
    xmin=0,
    xmax=51,
    ymin=0,
    ymax=11,
    xtick={0,10,20,30,40,50,60,70,80,90},
    ytick={0,2,4,6,8,10,12},
    grid=major,
    tick align=outside,
    label style={font=\small},
    tick label style={font=\small},
]

\addplot[fill=red, draw=none, fill opacity=0.12, forget plot]
    coordinates {(4,0) (4,11) (12,11) (12,0)} \closedcycle;
\addplot[fill=red, draw=none, fill opacity=0.12, forget plot]
    coordinates {(28,0) (28,11) (43,11) (43,0)} \closedcycle;

\addplot[
    thick,
    blue,
    mark=*,
    mark size=1.2pt,
] table[
    x expr=\thisrow{reveal_time_s}-1,
    y=speed_mps,
    col sep=comma,
    restrict expr to domain={\thisrow{stage}}{1:1},
] {figures/comfort_and_speed/data2/simlingo24781.csv};

\end{axis}

\begin{axis}[
    title={Cumulative Mean Comfort},
    width=\linewidth,
    height=0.58\linewidth,
    axis y line*=right,
    axis x line=none,
    xlabel={},
    ylabel={Comfort},
    xmin=0,
    xmax=51,
    ymin=0,
    ymax=1.1,
    ytick={0.0,0.2,0.4,0.6,0.8,1.0},
    tick align=outside,
    label style={font=\small},
    tick label style={font=\small},
    legend style={
        at={(0.96,0.98)},
        legend columns=1,
        anchor=north east,
        font=\tiny,
        inner sep=1.5pt,
        row sep=-1.5pt,
    },
]

\addlegendimage{thick, blue, mark=*, mark size=1.2pt}
\addlegendentry{Ego Speed}

\addlegendimage{area legend, fill=red, fill opacity=0.12, draw=none}
\addlegendentry{Stationary (speed $<$ 0.1)}

\addplot[
    thick,
    red,
    const plot,
] table[
    x expr=\thisrow{reveal_time_s}-1,
    y=cumulative_comfort,
    col sep=comma,
    restrict expr to domain={\thisrow{stage}}{1:1},
] {figures/comfort_and_speed/data2/simlingo24781.csv};
\addlegendentry{Traditional Comfort}

\addplot[
    thick,
    orange,
    const plot,
] table[
    x expr=\thisrow{reveal_time_s}-1,
    y=cumulative_comfort,
    col sep=comma,
    restrict expr to domain={\thisrow{stage}}{3:3},
] {figures/comfort_and_speed/data2/simlingo24781.csv};
\addlegendentry{Fixed Implementation}

\addplot[
    thick,
    green!50!black,
    const plot,
] table[
    x expr=\thisrow{reveal_time_s}-1,
    y=cumulative_comfort,
    col sep=comma,
    restrict expr to domain={\thisrow{stage}}{4:4},
] {figures/comfort_and_speed/data2/simlingo24781.csv};
\addlegendentry{Distance-based Segmentation}

\end{axis}
\end{tikzpicture}
}
    \caption{Sample route of SimLingo~\cite{renz2025simlingo}, showing driving speed over time together with the cumulative comfort scores of the different implementations. We re-project our distance-based implementation on the time axis. The traditional metric considers driving comfortable when and only when the ego vehicle is standing still, marked by the red shaded bands. Even the slight movement from seconds 35 to 38 is deemed uncomfortable. When we apply our fixes, the metric correctly scores segments when the vehicle is driving comfortably, but still incentivizes standing still. Further applying our distance-based segmentation fixes both problems, not changing the comfort score when the ego vehicle is stationary.}
    \label{fig:comfort_and_speed}
\end{figure}
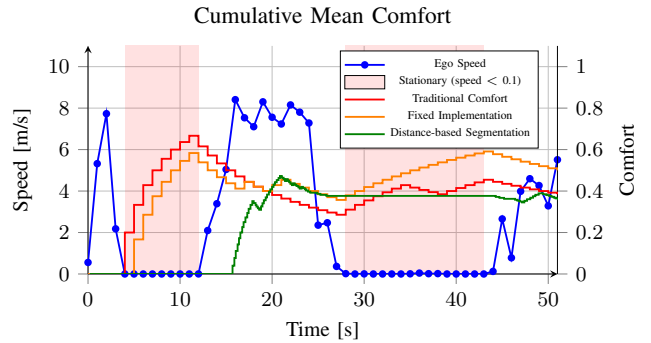

\Cref{fig:comfort_and_speed} shows the three variants on one SimLingo route, with our distance-based segments projected onto the time axis. The traditional metric rises only during the stationary bands and falls whenever the vehicle moves, even for the slight motion between seconds 35 and 38. The fixed implementation credits genuine driving but remains dominated by the stationary stretches, while ours is flat wherever the vehicle is stopped.

The three comfort columns of \cref{tab:comfort_main} let us state this progression as a number. Stationary time explains \SI{93}{\percent} of the variance in the traditional score ($r=0.97$), \SI{43}{\percent} of it after the implementation and smoothing fixes ($r=0.66$), and none of it under distance-based segmentation ($r=0.06$, $p=0.88$). Removing that dependence is precisely what distance-based segmentation was designed to do, so we read this as confirmation that the correction behaves as intended rather than as independent evidence that it measures comfort better; \cref{sec:efficiency_validation} supplies the external check.

\subsection{A Sanity Check Against Efficiency}
\label{sec:efficiency_validation}

Driving faster should intuitively be less comfortable. Bench2Drive measures speed relative to surrounding traffic as efficiency, computed independently of any comfort calculation, which lets us check cheaply whether a comfort metric behaves the way this intuition predicts. With eight models the correlations below are indicative rather than conclusive.

The traditional metric shows no such relationship ($r=-0.41$, $p=0.31$), while our distance-based metric does ($r=-0.69$, $p=0.054$; Spearman $\rho=-0.76$, $p=0.03$; $r=-0.68$ with the most efficient model, PDM-Lite, excluded). \textbf{Models that drive faster relative to surrounding traffic ride measurably less comfortably.} This tradeoff is invisible under the traditional metric, whose score is dominated by how long each model stood still.

\begin{figure}[t]
    \centering
    \definecolor{corrTrad}{HTML}{D55E00}
\definecolor{corrOurs}{HTML}{0072B2}

\pgfplotsset{
  corrpanel/.style={
    width=0.44\columnwidth,
    height=0.40\columnwidth,
    scale only axis,
    ylabel={comfort},
    grid=major, grid style={gray!18, line width=0.2pt},
    tick label style={font=\scriptsize},
    label style={font=\scriptsize},
    legend style={font=\tiny, draw=gray!60, fill=white,
                  fill opacity=0.92, text opacity=1, inner sep=2pt},
    legend cell align={left},
    tick align=outside,
  },
}

\resizebox{\linewidth}{!}{
\begin{tikzpicture}

  \begin{axis}[
    corrpanel,
    name=effPanel,
    xlabel={efficiency},
    xmin=120, xmax=265, ymin=0, ymax=90,
    legend style={at={(0.03,0.03)}, anchor=south west},
  ]
    \addplot[corrTrad, thick, no marks, forget plot]
      coordinates {(135.20,32.48) (250.80,17.68)};
    \addplot[only marks, mark=o, mark size=1.5pt,
             mark options={fill=corrTrad, draw=corrTrad}]
      table[x=efficiency, y=traditional, col sep=comma]
      {figures/comfort_efficiency_corr/corr_data.csv};
    \addlegendentry{Traditional ($r=-0.41$, $p=0.31$)}

    \addplot[corrOurs, thick, no marks, forget plot]
      coordinates {(135.20,68.79) (250.80,28.28)};
    \addplot[only marks, mark=triangle*, mark size=1.9pt,
             mark options={fill=corrOurs, draw=corrOurs}]
      table[x=efficiency, y=ours, col sep=comma]
      {figures/comfort_efficiency_corr/corr_data.csv};
    \addlegendentry{Ours ($r=-0.69$, $p=0.05$)}
  \end{axis}

  \begin{axis}[
    corrpanel,
    at={(effPanel.south east)}, xshift=1.9cm, anchor=south west,
    xlabel={driving score $-$ speed prediction},
    ylabel={comfort $-$ speed prediction},
    xmin=-19, xmax=22, ymin=-30, ymax=24,
    legend style={at={(0.97,0.03)}, anchor=south east},
  ]
    \addplot[gray!45, line width=0.2pt, no marks, forget plot]
      coordinates {(0,-30) (0,24)};
    \addplot[gray!45, line width=0.2pt, no marks, forget plot]
      coordinates {(-19,0) (22,0)};

    \addplot[corrTrad, thick, no marks, forget plot]
      coordinates {(-13.52,10.68) (16.66,-13.17)};
    \addplot[only marks, mark=o, mark size=1.5pt,
             mark options={fill=corrTrad, draw=corrTrad}]
      table[x=adj_ds, y=adj_traditional, col sep=comma]
      {figures/comfort_efficiency_corr/corr_data.csv};
    \addlegendentry{Traditional ($r=-0.80$, $p=0.03$)}

    \addplot[corrOurs, thick, no marks, dashed, forget plot]
      coordinates {(-13.52,-7.85) (16.66,9.68)};
    \addplot[only marks, mark=triangle*, mark size=1.9pt,
             mark options={fill=corrOurs, draw=corrOurs}]
      table[x=adj_ds, y=adj_ours, col sep=comma]
      {figures/comfort_efficiency_corr/corr_data.csv};
    \addlegendentry{Ours ($r=+0.44$, $p=0.32$, n.s.)}
  \end{axis}

\end{tikzpicture}
}
    \caption{Traditional comfort (vermillion circles) and our distance-based comfort (blue triangles), one point per model ($n=8$), least-squares fit per series. Left: comfort against efficiency; ours recovers the negative relationship the traditional metric misses. Right: the same against driving score with speed removed from both axes, each value being a model's score minus what its efficiency alone predicts, so zero means exactly as expected for that speed. At matched speed the traditional metric stays strongly negative, as expected of a score that tracks stationary time, while ours leaves no relationship; ours is dashed because at $n=8$ a tilt that size appears by chance about one time in three ($p=0.32$).}
    \label{fig:comfort_efficiency_corr}
\end{figure}
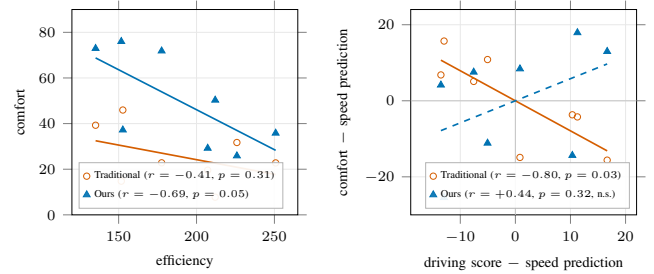

The corrected metric also changes what the benchmark says about driving score. Under the traditional metric, comfort and driving score are strongly negatively correlated ($r=-0.73$, $p=0.04$), which reads as though scoring well on Bench2Drive's edge cases requires harsh driving. Our metric weakens this to $r=-0.42$ ($p=0.31$), but that raw figure is still confounded: efficiency is not a term in the driving score, yet the two are strongly correlated in practice ($r=+0.85$, $p=0.007$), because models that stall or crawl leave routes incomplete and are terminated for being blocked. Comparing models at matched speed separates the two cases. Under our metric nothing remains: the fitted slope tilts slightly upward ($r=+0.45$), but with eight models a tilt that size arises by chance roughly one time in three ($p=0.32$), so we read it as no relationship rather than as comfort improving with driving score. Under the traditional metric the relationship stays strongly negative ($r=-0.80$), and there a tilt that size would arise by chance only about three times in a hundred ($p=0.03$), which is what a score that mostly tracks stationary time should do, since a stuck model both stands still and completes little of its route. \textbf{At matched driving speed we find no sign of the strong tradeoff the original metric reports, though eight models cannot rule out a weaker one.} \Cref{fig:comfort_efficiency_corr} shows both panels.

\subsection{Cross-Model Insights}
\label{sec:cross_model_insights}

\Cref{tab:comfort_main} spans several planning paradigms, and because \cref{sec:efficiency_validation} identified speed as the dominant confound we compare models by \emph{speed-adjusted} comfort, their residual from the least-squares fit of corrected comfort on efficiency. With one or two models per paradigm these are observations rather than controlled comparisons, but none of them are visible under the traditional metric, which orders these models by how long they stood still. The two language-model planners land at opposite extremes, Orion highest of the eight (75.98) and SimLingo lowest (25.86), separated by roughly 24 points even after adjusting for speed; because the two differ in how the plan leaves the language model, ride quality appears to be governed by the trajectory decoder rather than by the presence of a language model. The two pairs that happen to drive at matched speed point the same way: Orion and VAD sit within one efficiency point and Orion leads on both axes, while SparseDriveV2 and TFv6 sit within five and trade one against the other, SparseDriveV2 being 21.1 points more comfortable while scoring 7.3 points lower. CaRL, the only reinforcement-learned policy in our set, has the highest speed-adjusted comfort of the eight ($+17.9$) despite carrying no comfort term in its reward. Comfort is thus neither a byproduct of driving competence nor strictly opposed to it once speed is held fixed.

\section{CONCLUSIONS}
We examined Bench2Drive, the dominant closed-loop benchmark for end-to-end driving, and found three pitfalls in its evaluation protocol. The most consequential is its comfort metric, which measures little more than how long a vehicle stands still and tracks stationary time at $r=0.97$. Our corrected, distance-based metric cuts that correlation to $r=0.06$, and it changes what the benchmark says about the models scored under it. The strong tradeoff between driving score and comfort that prior evaluations report is no longer detectable once models are compared at matched driving speed, so comfort appears to trade against how fast a model drives rather than against how well it drives. Patterns the original metric could not surface also become visible. The two language-model planners we evaluate land at opposite extremes of the corrected scale despite sharing that design, which points at the trajectory decoder rather than the language model as what governs ride quality. The only reinforcement-learned policy in our set rides most smoothly for the speed it drives, without having been trained for comfort at all. We also identify two ways the harness itself distorts the navigation signal. The default controller can replace a model's own steering with an angle toward the route target point, and that target point can fall far outside the distribution a model was trained on. Both affect UniAD and VAD by very different amounts, so scores obtained under the default protocol are not comparable across models. Future closed-loop evaluations should therefore check for both before comparing models.

\section*{ACKNOWLEDGMENT}

The authors acknowledge the use of Claude (Anthropic) for drafting and refining text in results section and captioning figures, as well as GitHub Copilot for generating portions of the code used for prototyping, data processing and analysis. All AI-generated content was reviewed and verified by the authors.

\bibliographystyle{IEEEtran}
\bibliography{references}

\end{document}